\documentclass{article}
\makeatletter

\usepackage[verbose=true,letterpaper]{geometry}
\AtBeginDocument{
  \newgeometry{
    textheight=9in,
    textwidth=6.5in,
    top=1in,
    headheight=14pt,
    headsep=25pt,
    footskip=30pt
  }
}

\newcommand{\headeright}{A Preprint}
\newcommand{\undertitle}{A Preprint}
\newcommand{\shorttitle}{\@title}

\usepackage{fancyhdr}
\fancyheadoffset{0pt}
\def\keywordname{{\bfseries \emph{Keywords}}}%
\def\keywords#1{\par\addvspace\medskipamount{\rightskip=0pt plus1cm
\def\and{\ifhmode\unskip\nobreak\fi\ $\cdot$
}\noindent\keywordname\enspace\ignorespaces#1\par}}

\renewcommand{\normalsize}{%
  \@setfontsize\normalsize\@xpt\@xipt
  \abovedisplayskip      7\p@ \@plus 2\p@ \@minus 5\p@
  \abovedisplayshortskip \z@ \@plus 3\p@
  \belowdisplayskip      \abovedisplayskip
  \belowdisplayshortskip 4\p@ \@plus 3\p@ \@minus 3\p@
}
\normalsize
\renewcommand{\small}{%
  \@setfontsize\small\@ixpt\@xpt
  \abovedisplayskip      6\p@ \@plus 1.5\p@ \@minus 4\p@
  \abovedisplayshortskip \z@  \@plus 2\p@
  \belowdisplayskip      \abovedisplayskip
  \belowdisplayshortskip 3\p@ \@plus 2\p@   \@minus 2\p@
}
\renewcommand{\footnotesize}{\@setfontsize\footnotesize\@ixpt\@xpt}
\renewcommand{\scriptsize}{\@setfontsize\scriptsize\@viipt\@viiipt}
\renewcommand{\tiny}{\@setfontsize\tiny\@vipt\@viipt}
\renewcommand{\large}{\@setfontsize\large\@xiipt{14}}
\renewcommand{\Large}{\@setfontsize\Large\@xivpt{16}}
\renewcommand{\LARGE}{\@setfontsize\LARGE\@xviipt{20}}
\renewcommand{\huge}{\@setfontsize\huge\@xxpt{23}}
\renewcommand{\Huge}{\@setfontsize\Huge\@xxvpt{28}}

\providecommand{\section}{}
\renewcommand{\section}{%
  \@startsection{section}{1}{\z@}%
                {-2.0ex \@plus -0.5ex \@minus -0.2ex}%
                { 1.5ex \@plus  0.3ex \@minus  0.2ex}%
                {\large\bf\raggedright}%
}
\providecommand{\subsection}{}
\renewcommand{\subsection}{%
  \@startsection{subsection}{2}{\z@}%
                {-1.8ex \@plus -0.5ex \@minus -0.2ex}%
                { 0.8ex \@plus  0.2ex}%
                {\normalsize\bf\raggedright}%
}
\providecommand{\subsubsection}{}
\renewcommand{\subsubsection}{%
  \@startsection{subsubsection}{3}{\z@}%
                {-1.5ex \@plus -0.5ex \@minus -0.2ex}%
                { 0.5ex \@plus  0.2ex}%
                {\normalsize\bf\raggedright}%
}
\providecommand{\paragraph}{}
\renewcommand{\paragraph}{%
  \@startsection{paragraph}{4}{\z@}%
                {1.5ex \@plus 0.5ex \@minus 0.2ex}%
                {-1em}%
                {\normalsize\bf}%
}
\providecommand{\subparagraph}{}
\renewcommand{\subparagraph}{%
  \@startsection{subparagraph}{5}{\z@}%
                {1.5ex \@plus 0.5ex \@minus 0.2ex}%
                {-1em}%
                {\normalsize\bf}%
}

\newlength{\@abovecaptionskip}
\newlength{\@belowcaptionskip}

\renewenvironment{table}
  {\setlength{\abovecaptionskip}{\@belowcaptionskip}%
   \setlength{\belowcaptionskip}{\@abovecaptionskip}%
   \@float{table}}
  {\end@float}

\renewcommand{\footnoterule}{\kern-3\p@ \hrule width 12pc \kern 2.6\p@}
\def\@listi  {\leftmargin\leftmargini}
\def\@listii {\leftmargin\leftmarginii
              \labelwidth\leftmarginii
              \advance\labelwidth-\labelsep
              \topsep  2\p@ \@plus 1\p@    \@minus 0.5\p@
              \parsep  1\p@ \@plus 0.5\p@ \@minus 0.5\p@
              \itemsep \parsep}
\def\@listiii{\leftmargin\leftmarginiii
              \labelwidth\leftmarginiii
              \advance\labelwidth-\labelsep
              \topsep    1\p@ \@plus 0.5\p@ \@minus 0.5\p@
              \parsep    \z@
              \partopsep 0.5\p@ \@plus 0\p@ \@minus 0.5\p@
              \itemsep \topsep}
\def\@listiv {\leftmargin\leftmarginiv
              \labelwidth\leftmarginiv
              \advance\labelwidth-\labelsep}
\def\@listv  {\leftmargin\leftmarginv
              \labelwidth\leftmarginv
              \advance\labelwidth-\labelsep}
\def\@listvi {\leftmargin\leftmarginvi
              \labelwidth\leftmarginvi
              \advance\labelwidth-\labelsep}

\providecommand{\maketitle}{}
\renewcommand{\maketitle}{%
  \par
  \begingroup
    \renewcommand{\thefootnote}{\fnsymbol{footnote}}
    \long\def\@makefntext##1{%
      \parindent 1em\noindent
      \hbox to 1.8em{\hss $\m@th ^{\@thefnmark}$}##1
    }
    \thispagestyle{empty}
    \@maketitle
    \@thanks
  \endgroup
  \let\maketitle\relax
  \let\thanks\relax
}

\newcommand{\@toptitlebar}{
  \hrule height 2\p@
  \vskip 0.25in
  \vskip -\parskip%
}
\newcommand{\@bottomtitlebar}{
  \vskip 0.29in
  \vskip -\parskip
  \hrule height 2\p@
  \vskip 0.09in%
}

\providecommand{\@maketitle}{}
\renewcommand{\@maketitle}{%
  \vbox{%
    \hsize\textwidth
    \linewidth\hsize
    \vskip 0.1in
    \@toptitlebar
    \centering
    {\LARGE\sc \@title\par}
    \@bottomtitlebar
    \textsc{\undertitle}\\
    \vskip 0.1in
    \def\And{%
      \end{tabular}\hfil\linebreak[0]\hfil%
      \begin{tabular}[t]{c}\bf\rule{\z@}{24\p@}\ignorespaces%
    }
    \def\AND{%
      \end{tabular}\hfil\linebreak[4]\hfil%
      \begin{tabular}[t]{c}\bf\rule{\z@}{24\p@}\ignorespaces%
    }
    \begin{tabular}[t]{c}\bf\rule{\z@}{24\p@}\@author\end{tabular}%
  \vskip 0.4in \@minus 0.1in \center{\@date}   \vskip 0.2in
  }
}

\newcommand{\ftype@noticebox}{8}
\newcommand{\@notice}{%
  \enlargethispage{2\baselineskip}%
  \@float{noticebox}[b]%
    \footnotesize\@noticestring%
  \end@float%
}

\renewenvironment{abstract}
{
  \centerline
  {\large \bfseries \scshape Abstract}
  \begin{quote}
}
{
  \end{quote}
}

\makeatother

\usepackage[utf8]{inputenc}
\usepackage[T1]{fontenc}
\usepackage{amsmath,amssymb,booktabs,array,graphicx}
\usepackage[section]{placeins}
\newcolumntype{P}[1]{>{\raggedright\arraybackslash}p{#1}}
\usepackage{microtype}
\usepackage[numbers,sort&compress]{natbib}
\usepackage[hidelinks]{hyperref}
\usepackage{url}
\usepackage{listings}
\lstdefinestyle{studyjson}{
  basicstyle=\ttfamily\footnotesize,
  columns=fullflexible,keepspaces=true,
  showstringspaces=false,breaklines=true,
  frame=single,framerule=0.2pt,
  xleftmargin=0.5em,xrightmargin=0.5em,
  aboveskip=0.5em,belowskip=0.5em,
  captionpos=t
}
\renewcommand{\shorttitle}{Do System One Decisions Add Up?}
\renewcommand{\headeright}{Research preprint}
\renewcommand{\undertitle}{}
\title{Do System One Decisions Add Up?\\A Study of Probabilistic Coherence}
\author{Saman Sarker Joy\\
\normalfont Faculty of Computer Science and Information Technology\\
\normalfont Universiti Malaya\\
\normalfont Kuala Lumpur, Malaysia}
\date{September 27, 2026}
\hypersetup{pdftitle={Do System One Decisions Add Up? A Study of Probabilistic Coherence},pdfauthor={Saman Sarker Joy},pdfsubject={Jev and Laya: full empirical study},pdfkeywords={decision models, probabilistic coherence, hierarchical classification, calibration}}
\newcommand{\TV}{\operatorname{TV}}
\newcommand{\JSD}{\operatorname{JSD}}
\newcommand{\CFCE}{\operatorname{CFCE}}
\newcommand{\argmax}{\operatorname*{arg\,max}}
\newcommand{\ind}{\mathbf{1}}
\begin{document}
\maketitle
\begin{abstract}
A decision model can give probabilities that sum to one for every question yet disagree with itself when the same decision is broken into smaller steps. We study this form of probabilistic coherence in Jev and the English Laya checkpoint, using 2,500 matched examples per system across TREC, CLINC150, and MASSIVE. Across 72,000 classification questions, we compare direct fine-label predictions with broad-category probabilities and predictions reconstructed through those categories. Both systems show substantial disagreement: mean category-level total variation ranges from 0.219 to 0.349 for Jev and from 0.424 to 0.689 for Laya, on a scale where zero means exact agreement. The consequences differ sharply. On CLINC150, reconstruction reduces Jev's accuracy by 22.9 percentage points (paired 95\% bootstrap interval: $[-24.9,-20.9]$) and improves Laya's by 21.3 points ($[18.0,24.5]$). The same directions hold across all three datasets, with all six unadjusted accuracy-change intervals excluding zero. Improved accuracy can also accompany less reliable confidence: on MASSIVE, Laya gains 9.2 accuracy points while its expected calibration error rises from 0.046 to 0.124. Error analysis identifies both broad-category mistakes and within-category confusions. These findings show why decision systems need joint evaluation of accuracy, confidence calibration, and probability coherence in the workflow used by an application.
\end{abstract}

\section{Introduction}
A natural-language decision can often be represented in more than one way. A customer request may be classified directly into a specific intent, or first assigned to a domain and then classified among that domain's intents. A factual question may likewise be categorized by its broad answer type before a narrower subtype is selected. These choices affect deployment: the resulting probabilities can determine whether an application routes a case automatically, requests additional information, or escalates to a human. An application that decomposes a decision therefore needs to know whether the model's probability assignments remain compatible across the decomposition.

This question is particularly relevant for structured decision interfaces, which return a choice and probabilities over the available alternatives. TypeSafe calls this approach System One \citep{typesafe}. We use that term for the interface style; the underlying model architectures can differ. Jev and Laya both support this kind of classification \citep{typesafe,laya}. Their usefulness depends on selecting the right label and providing probabilities that an application can meaningfully combine.

Accuracy, calibration, and coherence answer three different questions. Accuracy asks whether the selected label is correct. Calibration asks whether confidence matches observed success: among predictions made with 80\% confidence, roughly 80\% should be correct \citep{guo}. Coherence asks whether probabilities for related events agree. A model may choose the correct label under two formulations while assigning very different probabilities to it. It may also produce mutually consistent probabilities while repeatedly choosing the wrong label. Each property therefore needs its own evaluation.

Probability theory gives two useful checks. First, a broad category's probability should equal the sum of the probabilities of its fine labels. Second, a fine label's probability should equal its parent's probability multiplied by the label's probability conditional on that parent. We test how closely separately asked questions satisfy these relationships. Asking a model to assume a category changes both its instructions and its available choices, so the answers may reflect different interpretations. Our use of \emph{cross-formulation coherence} concerns the practical compatibility of these outputs. Claims about a model's internal beliefs would require additional evidence.

The study uses three established classification datasets with explicit two-level taxonomies: question classification in TREC \citep{liroth}, intent classification in CLINC150 \citep{larson}, and English intent classification in MASSIVE \citep{fitzgerald}. The paired design holds examples and semantic label identities fixed while comparing direct and decomposed decisions. Importantly, it queries the children of \emph{every} parent. This reconstructs unconditional probabilities over the complete fine-label space.

This study contributes an auditable comparison of three decision rules built from complete categorical outputs; paired evidence that decomposition changes accuracy in opposite directions for the two tested systems; and diagnostics linking those changes to routing errors, confidence, and presentation sensitivity. The contribution combines an empirical evaluation protocol with a detailed two-system case study.

Code and recorded experimental results are available in the project repository.\footnote{\url{https://github.com/samanjoy2/system-one-coherence}}

We address four research questions:
\begin{description}
\item[\normalfont RQ1:] Do broad-category probabilities agree with the summed probabilities of their fine labels?
\item[\normalfont RQ2:] Do probabilities reconstructed through broad categories agree with direct fine-label probabilities?
\item[\normalfont RQ3:] How does breaking a decision into stages affect prediction accuracy and confidence calibration?
\item[\normalfont RQ4:] How stable are the findings under reordered choices, repeated requests, ties, and different calibration-bin counts?
\end{description}
\section{Related work}
\subsection{Probabilistic consistency of language-model judgments}
Previous work treats probabilistic coherence as a measurable property of language-model judgments. Betz and Richardson \citep{betz} study coherence, logical consistency, and belief updating in a controlled artificial-language setting. Their work motivates evaluating consistency alongside task success. We examine released decision systems on existing natural-language classification benchmarks.

Wolf et al.\ \citep{wolf} provide a closely related partition-and-aggregation approach. They compare direct population estimates with estimates combined from prompted subpopulations. We apply the same general consistency principle to a fixed hierarchy of mutually exclusive labels for one input. Our contribution is an evaluation of categorical decision interfaces that connects probability disagreement to classification accuracy, calibration, and routing errors.

Consistency constraints also appear in computer vision: Dual-Align uses topology-preserving and cycle-consistency losses to align masked and clean face representations \citep{anwar2026dualalign}.

\subsection{Calibration and hierarchical decision making}
Confidence and accuracy can change independently. For example, temperature scaling adjusts probabilities while preserving the highest-scoring class \citep{guo}. We retain each system's released calibration settings, so the evaluation includes their effects. This matters particularly for Laya, whose temperature depends on the number of available choices.

Confidence also has limits as a guide to the usefulness of supervision. In low-resource summarization, Sumit et al.\ \citep{sumit2026doesknowledgedistillationhurt} find weak associations between teacher-confidence measures and distillation usefulness estimated through gradient alignment. Their findings motivate evaluating confidence against a task-specific outcome. Our study examines a different outcome: the reliability and coherence of probabilities combined across classification questions.

A label hierarchy provides two ways to make a staged decision. Hard routing chooses one parent and then one of its children. Marginalization keeps every parent in consideration and combines probabilities across branches. These rules can select different labels even with perfectly coherent probabilities: the strongest individual child may belong to the second-ranked parent. Our coherence comparison therefore uses direct fine-label probabilities and probabilities reconstructed from the hierarchy.

\subsection{Sensitivity to candidate presentation}
Candidate order and identifiers can affect multiple-choice model behavior. Zheng et al.\ \citep{zheng} document option-selection bias in generative language models. This motivates checking presentation sensitivity in decision systems as well. We keep each label's identifier fixed, change its display position, and evaluate the released systems without additional debiasing.

Our protocol uses the interface's complete probability distribution over the supplied alternatives. Having every candidate's probability allows direct comparisons and reconstruction across the hierarchy.

\section{Problem formulation}
\label{sec:formulation}
Each dataset groups specific labels into broad categories, which we call parents. Let $x_i$ denote an input and $y_i\in\mathcal C$ its correct fine label. The set of parents is $\mathcal P$, and the mapping $g:\mathcal C\rightarrow\mathcal P$ assigns every fine label to exactly one parent. The children of parent $k$ are $\mathcal C_k=\{c:g(c)=k\}$. There are $K=|\mathcal P|$ parents and $L=|\mathcal C|$ fine labels. Every parent has at least one child, and the child sets are disjoint and cover all candidate labels. The correct parent is $g(y_i)$. For readability, the equations below omit the input and model indices when these are fixed.

We ask three kinds of question: choose a broad category, choose directly among all fine labels, and choose a fine label assuming a specified parent. Their probability distributions are
\begin{align}
u(k)&=\widehat P_{\mathrm{coarse}}(k\mid x), && k\in\mathcal P,\\
p(c)&=\widehat P_{\mathrm{flat}}(c\mid x), && c\in\mathcal C,\\
v_k(c)&=\widehat P_{\mathrm{conditional}}(c\mid x,\text{assume }k),
&&c\in\mathcal C_k.
\end{align}
Each distribution is checked and normalized over its own choices. The hat marks a probability returned by the model. Since $v_k$ comes from a separate question, its relationship to $p$ is an empirical question that the study tests.

\subsection{Three decision structures}
Flat classification takes the most probable label from the direct question:
\begin{equation}
\widehat y_{\mathrm{flat}}=\argmax_{c\in\mathcal C}p(c).
\end{equation}
Hard routing commits to the most probable parent, $\widehat k=\argmax_k u(k)$, and then chooses its most probable child:
\begin{equation}
\widehat y_{\mathrm{hard}}=\argmax_{c\in\mathcal C_{\widehat k}}v_{\widehat k}(c).
\end{equation}
A wrong parent decision excludes the correct child from this route. Hard routing supplies a selected label; probability metrics use the complete flat and marginal vectors.

Full marginalization keeps all parent branches. It multiplies each fine label's within-parent probability by its parent's probability, then selects the largest resulting value:
\begin{equation}
q(c)=u\!\left(g(c)\right)v_{g(c)}(c),\qquad
\widehat y_{\mathrm{marginal}}=\argmax_{c\in\mathcal C}q(c).
\label{eq:q}
\end{equation}
Because each child has exactly one parent, the usual sum over parents reduces to one nonzero contribution for each child. Nevertheless, all parent branches must be evaluated to construct the entire vector. Normalization follows directly:
\begin{equation}
\sum_{c\in\mathcal C}q(c)
=\sum_{k\in\mathcal P}u(k)\sum_{c\in\mathcal C_k}v_k(c)
=\sum_{k\in\mathcal P}u(k)=1.
\end{equation}
Ties in all argmax operations are resolved by canonical lexicographic label order.

For illustration only, consider two parents with masses $0.55$ and $0.45$. If the highest conditional child probabilities in their respective branches are $0.51$ and $0.90$, hard routing chooses a child in the first branch, but the second branch contains a child with greater joint mass ($0.405$ versus $0.2805$). This invented arithmetic example illustrates the distinction between marginalization and hard routing.

\subsection{Compatibility conditions and their limits}
For RQ1, we sum the direct fine-label probabilities within each parent:
\begin{equation}
a(k)=\sum_{c\in\mathcal C_k}p(c).
\end{equation}
Exact category-level coherence means $u=a$. For RQ2, exact fine-label coherence means $p=q$. Fine-label agreement guarantees parent agreement, while equal parent totals can conceal differences among their children. Summing $q$ within a parent always recovers $u$, which provides a check on the implementation. The following inequality makes the relationship precise: parent-level disagreement is a lower bound on fine-label disagreement.
\begin{equation}
\underbrace{\frac12\sum_k|a(k)-u(k)|}_{\text{parent discrepancy}}
=\frac12\sum_k\left|\sum_{c\in\mathcal C_k}(p(c)-q(c))\right|
\leq\underbrace{\frac12\sum_c|p(c)-q(c)|}_{\text{fine-label discrepancy}}.
\label{eq:contraction}
\end{equation}
Thus a parent-level mismatch certifies at least that much fine-label total variation, although a small parent mismatch need not imply fine-label agreement. This follows from the standard aggregation property of total variation.

These checks concern the probabilities returned under the tested question formulations. Ambiguous labels and changed instructions can affect their interpretation. The findings therefore describe the evaluated workflows; isolating the effect of hierarchy itself would require controlled changes to wording and candidate presentation.

\section{Methodology}
\label{sec:methods}
\subsection{Design and analysis units}
Both systems evaluate the same 2,500 inputs, giving 5,000 completed model--example records. For each input, we ask one broad-category question, one direct fine-label question, and one conditional question for each parent: $K+2$ questions in total. Across both systems, the evaluation contains 72,000 classification questions. Statistical comparisons use matched input examples as the unit of analysis. The total is 36,000 questions per system, excluding audits and retries.

The primary analysis unit is an example within a dataset and model configuration. Results are reported separately for each dataset. The three datasets differ in label count, hierarchy, sampling distribution, and domain. Within-model comparisons of decision structures use matched examples. Between-model contrasts describe the evaluated systems as deployed; architecture, parameter count, and precision vary jointly.

\subsection{Datasets, taxonomies, and sampling}
\begin{table}[t]
\caption{Fixed evaluation design. Represented labels are fine labels with at least one selected gold example; all declared labels remain candidates. Counts describe the fixed evaluation sample.}
\label{tab:data}
\centering
\small
\setlength{\tabcolsep}{5pt}
\begin{tabular}{@{}lrrrrp{0.24\linewidth}@{}}
\toprule
Dataset & Examples & Parents & Fine labels & Represented & Selection\\
\midrule
TREC & 500 & 6 & 50 & 42 & Entire test split\\
CLINC150 & 1,000 & 10 & 150 & 150 & Stratified in-scope test\\
MASSIVE en-US & 1,000 & 18 & 60 & 59 & Capacity-aware test sample\\
\bottomrule
\end{tabular}
\end{table}

\paragraph{TREC.}
We use all 500 test questions from the TREC question-classification dataset \citep{liroth}, retrieved from the pinned CogComp Parquet conversion. Its six broad classes are Abbreviation, Description, Entity, Human, Location, and Numeric, with 50 declared fine classes. Eight fine classes have no gold instances in the test split, but remain valid candidates. Coarse descriptions expand the broad abbreviations; fine descriptions join the expanded parent name with the original fine-label suffix, such as \texttt{Entity: cremat}. The suffixes are not rewritten into explanatory prose. This choice preserves the implemented taxonomy but creates a label-interpretability limitation.

\paragraph{CLINC150.}
We use 1,000 examples from the in-scope test portion of CLINC150 \citep{larson}. The official domain mapping supplies ten domains, each containing 15 intents. Out-of-scope examples are excluded because the present experiment requires an exhaustive, closed-set fine-label partition. Each intent contributes six or seven selected examples. Candidate descriptions replace underscores with spaces without introducing demonstrations or hand-written intent definitions.

\paragraph{MASSIVE.}
We use the English \texttt{en-US} test partition of MASSIVE version 1.0 \citep{fitzgerald}. Official scenario and intent annotations define 18 parents and 60 fine labels. The taxonomy is constructed from the English release's annotations, while evaluation examples come only from its test partition. Sampling is capacity-aware because rare intents cannot meet a uniform quota. The \texttt{cooking\_query} intent has no test examples and remains a candidate; 59 fine labels are represented in the selected sample. Two pairs of duplicate input texts occur under distinct source identifiers and are retained as separate source records with their distinct source identities.

\paragraph{Sampling algorithm.}
For CLINC150 and MASSIVE, seed 42 controls a deterministic approximate class-balancing procedure. An initial quota is $\lfloor n/L\rfloor$ per fine class; the remainder is distributed using a seeded permutation. Each quota is capped at the number of available test examples. Any shortfall is redistributed one example at a time to an eligible class with the smallest current quota, using a seeded tie order. Within each class, source records are sorted by identifier and sampled without replacement. Selected records are then sorted by identifier and frozen.

For CLINC150 and MASSIVE, this defines an approximately label-balanced evaluation population subject to test-set capacity. TREC instead retains its original, imbalanced test composition. Calibration and average loss pertain to these respective evaluation mixtures; no importance weighting to a deployment distribution is applied. All systems and decision structures use the same frozen examples. Source identifiers, texts, labels, taxonomies, and source-file hashes are retained to detect accidental sample changes.

\subsection{Systems and serving configurations}
\begin{table}[!htbp]
\centering\small
\caption{Evaluated configurations. Neither system is fitted to the study examples. Architecture descriptions rely on release documentation; hosted internals remain unverified.}
\label{tab:models}
\begin{tabular}{P{.10\linewidth}P{.28\linewidth}P{.21\linewidth}P{.29\linewidth}}
\toprule
System & Identity & Execution & Qualification\\
\midrule
Jev & \texttt{jev-1.13.0} & TypeSafe API; precision undisclosed & Proprietary training and serving details\\
Laya & Pinned English root checkpoint & Local FP32; runtime 0.3.20 & Expanded context; released option-count calibration\\
\bottomrule
\end{tabular}
\end{table}

\paragraph{Jev.}
Jev is accessed through TypeSafe's native categorical Choice interface with version \texttt{jev-1.13.0} \citep{typesafe}. The study uses the interface's native categorical probabilities. Its training corpus, parameter count, and internal serving precision are not established by this evaluation. API probabilities are recorded at the precision returned by the service.

\paragraph{Laya.}
We evaluate only the English root \texttt{convaiinnovations/laya} checkpoint, pinned in the source lock, using runtime 0.3.20 \citep{laya}. The model card describes a ModernBERT-large-based system of approximately 421 million parameters. The local model is placed in evaluation mode and FP32, with automatic mixed precision disabled. No language routing, sibling checkpoint selection, or study-specific fine-tuning is performed.

The default total/head token budgets of 512/192 truncate large candidate sets in this protocol. The evaluation instead fixes them at 1,088/960, within the encoder's declared 8,192-token capacity. Every constructed input is checked against the expected complete token sequence; a truncated or rewritten sequence is rejected. The checkpoint's released calibration is retained as implemented by the runtime. In particular, the stored temperature for 11 or more choices, approximately 0.10058, is clamped by that runtime to 0.5. Temperatures for two, three--five, and six--ten choices are approximately 1.90636, 1.76015, and 1.00002. The evaluation preserves this released behavior. It can contribute to differences across candidate-set sizes. The model card also cautions about high-cardinality choices; the present evaluation includes such settings and should not be generalized to all intended Laya applications.

\paragraph{Scope.}
We evaluate Jev and the English Laya checkpoint on the same examples using identical question templates.

\subsection{Question construction and candidate controls}
The input text is passed unchanged as the state. The three instruction templates are:
\begin{quote}\small
Coarse: Classify the input into one of these broad categories.\\
Flat: Classify the input into one of these specific categories.\\
Conditional: Assuming the input belongs to \{parent description\}, classify it into one of these specific categories.
\end{quote}
Appendix~\ref{app:json} shows real JSON inputs and saved outputs for both systems, with a worked probability comparison.
The first question contains all parents; the second contains all fine labels; each conditional question contains only the children of its named parent. Ground-truth labels are not inserted into the state or instruction. Every parent is queried regardless of the coarse prediction or gold parent.

Each fine label keeps the same identifier in direct and conditional questions. Choices appear in a reproducible shuffled order, using the same procedure for both systems. Appendix~\ref{app:implementation} gives the identifier and ordering implementation.

Each conditional subset has its own shuffled order. The comparison therefore changes instructions, the number and identity of choices, and potentially their relative positions together. The order audit examines presentation sensitivity separately, while other sources of change remain combined in the main comparison.

\subsection{Checking and normalizing probabilities}
We check that every supplied candidate receives a finite probability in $[0,1]$. The probabilities must have a positive sum, $s$. Because the systems round their outputs, a small difference from one is allowed:
\begin{equation}
|s-1|\leq m\left(\tfrac12\,10^{-d}\right)+10^{-6},
\end{equation}
Here, $m$ is the number of choices and $d$ is the configured decimal precision: two for Jev and four for Laya. We divide accepted probabilities by $s$ so that they sum to one, while retaining the original responses. The bound allows accumulated rounding across choices; the actual deviations are checked separately below. Very small differences remain limited by the resolution of the recorded outputs. All reported examples have complete, validated answers. Execution and recovery details appear in Appendix~\ref{app:implementation}.

\subsection{Predictive performance and calibration}
For each dataset and system, accuracy and macro F1 are computed for flat, hard, and marginal predictions; coarse performance is evaluated against gold parents. Macro F1 averages over the \emph{entire declared} label set with undefined class F1 assigned zero, including classes with no gold support. Thus TREC uses all 50 fine classes and MASSIVE all 60, including absent classes. This convention must accompany any comparison with published macro-F1 scores.

For RQ3, we also evaluate the quality of the probabilities. The Brier score measures squared error across the full probability distribution, with lower values indicating better predictions \citep{gneiting}. For a distribution $r_i$ over the label set, it is
\begin{equation}
\operatorname{Brier}(r)=\frac1N\sum_{i=1}^{N}
\sum_{c}\left(r_i(c)-\ind[y_i=c]\right)^2.
\end{equation}
This summed score lies in $[0,2]$. Expected calibration error (ECE) measures the gap between confidence and accuracy. We group predictions into 15 equally spaced confidence bins and compute
\begin{equation}
\operatorname{ECE}_{15}(r)=
\sum_{b=1}^{15}\frac{|B_b|}{N}
\left|\operatorname{acc}(B_b)-\operatorname{conf}(B_b)\right|.
\end{equation}
Here, $B_b$ contains the examples in bin $b$, and confidence is $\max_c r_i(c)$. Bins include their lower endpoint and exclude their upper endpoint, except that the final bin includes one. Empty bins contribute zero. Lower ECE indicates closer agreement between confidence and accuracy at this bin resolution. Brier and ECE use the complete coarse, flat, and marginal distributions; hard routing supplies only the selected label for our comparisons.

\subsection{Cross-formulation coherence measures}
For RQ1, we compare asked parent probabilities $u$ with summed fine-label probabilities $a$. Coarse--fine coherence error (CFCE) averages their absolute differences; total variation (TV) rescales the discrepancy to $[0,1]$:
\begin{align}
\CFCE_i&=\frac1K\sum_{k\in\mathcal P}|u_i(k)-a_i(k)|,\\
\TV_i^{\mathrm{parent}}&=\frac12\sum_{k\in\mathcal P}|u_i(k)-a_i(k)|
=\frac K2\CFCE_i.
\end{align}
Zero indicates exact agreement for both measures. CFCE ranges from zero to $2/K$, so comparisons across datasets must account for the number of parents. TV has the common $[0,1]$ scale and is the main category-level measure used in our interpretation.

For RQ2, gold-label probability drift measures how much the probability of the correct label changes between direct and reconstructed predictions:
\begin{equation}
D_i=|p_i(y_i)-q_i(y_i)|.
\end{equation}
The correct label stays fixed in this comparison. To capture changes among all labels, we also compute Jensen--Shannon divergence (JSD), using the average distribution $m_i=(p_i+q_i)/2$:
\begin{equation}
\JSD_i=\tfrac12\operatorname{KL}_2(p_i\Vert m_i)
+\tfrac12\operatorname{KL}_2(q_i\Vert m_i).
\end{equation}
Here, $\operatorname{KL}_2$ is Kullback--Leibler divergence with base-two logarithms; zero-mass terms contribute zero. JSD \citep{lin} ranges from zero to one bit, with zero indicating identical distributions. Both JSD and parent TV can be calculated from model outputs and the taxonomy alone.

Finally, we report how often flat predictions select the same label as hard routing or marginalization. This complements the probability measures because the winning label can stay unchanged while its probability moves substantially. We summarize coherence using means, medians, sample standard deviations, and intervals for the means. Exploratory summaries also distinguish flat-correct, flat-incorrect, and flat-confidence-at-least-0.8 examples. These groups overlap and describe associations.

\subsection{Uncertainty estimation and comparisons}
We estimate uncertainty by repeatedly resampling the evaluated examples, using 10,000 bootstrap samples with seed 44. Sampling takes place within each correct fine-label group, preserving its size. The same resampled examples are used for every decision rule, which keeps comparisons paired. All metrics, including macro F1 and ECE, are recomputed for each sample. The middle 95\% of the resulting estimates, bounded by the 2.5th and 97.5th percentiles, gives the reported interval.

For comparisons with flat classification, we calculate the metric difference within every resampled set and form an interval from those paired differences. Broad-category metrics use the same fine-label sampling groups. Because broad and fine classification have different possible answers, their scores describe different tasks.

The intervals describe resampling variation for the observed examples and label composition. Training variation, service changes, and alternative taxonomies remain outside their coverage. A label represented by one example contributes no within-label variation, and duplicate texts can weaken the independence assumption. Intervals are unadjusted for multiple comparisons and support descriptive conclusions about these configurations. General rankings across model families require a broader design.

The bootstrap treats the observed examples as an empirical reference population with fixed label composition. In particular, TREC includes its entire released test split: its interval summarizes resampling sensitivity around the known finite-test accuracy.

\subsection{Robustness audits}
\paragraph{Option order.}
A fixed audit set contains 90 examples, 30 per dataset, selected with seed 43 and approximately balanced allocation across parents. Four additional candidate-order permutations are evaluated relative to the main permutation, using the same examples across systems. Comparisons align probabilities by semantic label before computing total variation, mean and maximum absolute changes, label flips, and exact-vector matches for coarse, flat, and reconstructed marginal distributions. Each dataset therefore contributes 120 permutation-to-main comparisons per system, but only 30 distinct examples. These repeated comparisons are not independent samples; current audit summaries are descriptive, and any future uncertainty analysis must cluster or resample by example.

\paragraph{Repeated requests.}
Jev receives two additional identical-request evaluations on the same audit examples, retaining input text, identifiers, and option order. These comparisons assess repeated-request variability separately from deliberate option reordering. Exact equality concerns the recorded, rounded vectors; it cannot establish equality of unrounded latent scores or determinism under all service conditions.

\paragraph{Exploratory error analysis.}
After the main analysis, we partition flat--marginal outcomes into both correct, lost correctness, gained correctness, and both wrong. Separately, hard-routing errors are partitioned into wrong-parent cases and correct-parent/wrong-child cases. We tabulate flat high-confidence errors, label-pair confusions, and subgroup JSD. These are post-hoc descriptive diagnostics. Illustrative examples are selected deterministically: the largest absolute gold-probability drift among Jev lost-correct cases or Laya gained-correct cases within each dataset, breaking ties by source identifier. Their selection emphasizes extreme behavior. No qualitative category frequencies are inferred from this selected set.

\paragraph{Post-hoc sensitivity checks.}
During manuscript review, we additionally recount exact maxima ties in recorded flat and marginal vectors, recompute ECE with 10 and 20 bins alongside the primary 15-bin specification, and recompute accuracy changes after retaining only the lexicographically first identifier for each duplicate input text. For ties, lower and upper accuracy bounds allow any label among the tied maxima while leaving all non-tied predictions fixed. These checks use existing records only, do not replace the primary protocol, and receive no new significance tests.

\subsection{Data integrity and completed coverage}
Dataset versions and selected examples are fixed throughout the comparison. All 5,000 primary records passed checks of input identity, label hierarchy, probability normalization, and recomputed metrics. The study also contains 720 order-audit records and 180 repeated-request records. Each of the 72,000 primary questions was checked against its saved input, original response, and normalized probabilities. Tables and figures use these verified records and the final 10,000-sample bootstrap analysis.

The cache check passed for all primary questions. Before normalization, 1,303 of Jev's 36,000 vectors and 15,062 of Laya's 36,000 vectors differ from unit mass by more than $10^{-8}$. Their maximum absolute mass errors are 0.0100 and 0.0017, respectively. These small observed residuals are distinct from the larger theoretical acceptance bounds. They validate the recorded normalization procedure, but cannot bound all effects of individual-entry rounding or recover unrounded probabilities.

These checks establish the integrity of the recorded evaluation. Jev's serving precision remains unknown, while Laya is evaluated in FP32. Evidence in this paper is restricted to the two completed configurations.

\section{Results}
\label{sec:results}
\subsection{Coverage and predictive performance}
Breaking classification into stages reduces Jev's accuracy and improves Laya's across all three datasets (RQ3). The analysis includes all 500 TREC, 1,000 CLINC150, and 1,000 MASSIVE examples for each system. Main results use the fixed primary choice order; the robustness audit examines alternative orders.

Table~\ref{tab:classification} reports all three fine-label decision rules. Jev's flat accuracies are 72.2\%, 91.0\%, and 83.3\% on TREC, CLINC150, and MASSIVE. Marginalization lowers these to 60.4\%, 68.1\%, and 66.9\%. Laya exhibits the opposite direction: flat accuracies of 28.8\%, 23.1\%, and 23.8\% increase to 34.4\%, 44.4\%, and 33.0\%. Jev remains more accurate than Laya for each matched fine-label rule in every dataset. This descriptive ordering applies to the evaluated configurations, whose architecture, scale, and training data differ.

\begin{table}[!htbp]
\centering\small
\setlength{\tabcolsep}{4pt}
\caption{Fine-label performance (percent). Macro F1 includes every declared class, including absent test classes. F, H, and M denote flat, hard routing, and marginalization.}
\label{tab:classification}
\begin{tabular}{llrrrrrr}
\toprule
Dataset & Model & F acc. & F F1 & H acc. & H F1 & M acc. & M F1\\
\midrule
TREC & Jev & 72.2 & 48.6 & 57.0 & 35.7 & 60.4 & 37.4\\
TREC & Laya & 28.8 & 24.7 & 33.4 & 31.7 & 34.4 & 32.9\\
CLINC150 & Jev & 91.0 & 90.4 & 64.6 & 62.3 & 68.1 & 66.1\\
CLINC150 & Laya & 23.1 & 28.3 & 33.9 & 33.9 & 44.4 & 44.6\\
MASSIVE & Jev & 83.3 & 81.1 & 65.9 & 65.1 & 66.9 & 66.0\\
MASSIVE & Laya & 23.8 & 24.8 & 32.9 & 30.4 & 33.0 & 30.3\\
\bottomrule
\end{tabular}
\end{table}

\begin{figure}[!htbp]
\centering
\includegraphics[width=\linewidth]{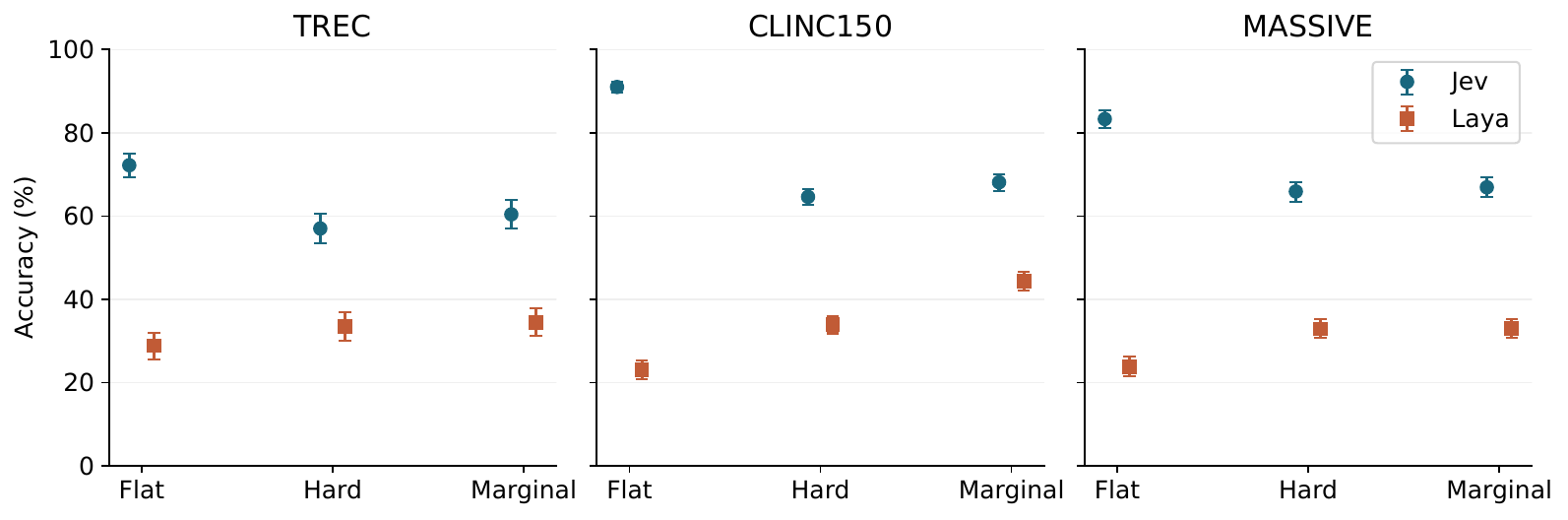}
\caption{Fine-label accuracy with 95\% stratified-bootstrap intervals. Both hierarchical rules reduce Jev's accuracy and increase Laya's relative to flat prediction. Marginalization retains mass from all branches through the full taxonomy.}
\label{fig:accuracy}
\end{figure}
The paired changes in Table~\ref{tab:accuracyci} range from $-22.9$ to $-11.8$ percentage points for Jev and from $+5.6$ to $+21.3$ points for Laya. Each interval excludes zero under the specified bootstrap; these are unadjusted descriptive intervals. Macro F1 follows the same direction as accuracy. The TREC gap between accuracy and macro F1 reflects uneven class support and the convention of including eight absent classes.

\begin{table}[!htbp]
\centering\small
\setlength{\tabcolsep}{4pt}
\caption{Accuracy and paired changes with 95\% stratified percentile-bootstrap intervals (10,000 replicates). Accuracies are percentages; changes are percentage points.}
\label{tab:accuracyci}
\begin{tabular}{llrrr}
\toprule
Dataset & Model & Flat & Marginal & Change\\
\midrule
TREC & Jev & 72.2 [69.2, 75.0] & 60.4 [57.0, 63.8] & -11.8 [-15.2, -8.4]\\
TREC & Laya & 28.8 [25.6, 32.0] & 34.4 [31.2, 37.8] & +5.6 [+1.4, +9.8]\\
CLINC150 & Jev & 91.0 [89.6, 92.3] & 68.1 [66.1, 70.0] & -22.9 [-24.9, -20.9]\\
CLINC150 & Laya & 23.1 [20.9, 25.4] & 44.4 [42.1, 46.7] & +21.3 [+18.0, +24.5]\\
MASSIVE & Jev & 83.3 [81.2, 85.3] & 66.9 [64.5, 69.3] & -16.4 [-18.6, -14.2]\\
MASSIVE & Laya & 23.8 [21.5, 26.2] & 33.0 [30.8, 35.2] & +9.2 [+6.3, +12.1]\\
\bottomrule
\end{tabular}
\end{table}

Full marginalization has higher observed accuracy than hard routing in all six model--dataset combinations. The gain varies substantially: Laya gains 10.5 points on CLINC150 (33.9\% to 44.4\%) and 0.1 points on MASSIVE (32.9\% to 33.0\%). These hard-to-marginal differences are descriptive; the reported paired intervals compare each rule with flat prediction.

Coarse accuracy reveals a different ordering from fine accuracy. On TREC, Laya's coarse accuracy is 89.2\%, versus Jev's 76.6\%, yet Laya's fine accuracy remains much lower. On CLINC150 and MASSIVE, Jev's coarse accuracies are 68.3\% and 72.9\%, versus Laya's 40.4\% and 47.7\%. A good broad classifier is therefore insufficient to establish a good fine-grained workflow.

\subsection{Accuracy gains and calibration changes diverge}
Marginalization worsens Jev's Brier score on all three datasets: from 0.401 to 0.560 on TREC, 0.136 to 0.440 on CLINC150, and 0.245 to 0.455 on MASSIVE. It improves Laya's Brier score from 0.928 to 0.742, 0.913 to 0.714, and 0.884 to 0.841 (Table~\ref{tab:calibration}). These changes agree in direction with the accuracy changes, and the paired Brier intervals exclude zero in all six cases.

Calibration gives a more mixed picture. Laya's MASSIVE accuracy improves by 9.2 points and Brier loss decreases, yet ECE rises from 0.046 to 0.124; its paired change is 0.078 with interval [0.044, 0.096]. On CLINC150, ECE decreases from 0.160 to 0.080. On TREC, the ECE-change intervals include zero for both systems. Thus an improvement in classification can accompany a larger confidence--accuracy gap.

\begin{table}[!htbp]
\centering\small
\setlength{\tabcolsep}{4pt}
\caption{Brier score (B) and 15-bin top-label ECE (E), lower is better. Brier uses the sum of classwise squared errors. Coarse scores concern a different label space.}
\label{tab:calibration}
\begin{tabular}{llrrrrrr}
\toprule
Dataset & Model & Coarse B & Coarse E & Flat B & Flat E & Marg. B & Marg. E\\
\midrule
TREC & Jev & 0.347 & 0.082 & 0.401 & 0.058 & 0.560 & 0.079\\
TREC & Laya & 0.175 & 0.033 & 0.928 & 0.203 & 0.742 & 0.169\\
CLINC150 & Jev & 0.463 & 0.137 & 0.136 & 0.020 & 0.440 & 0.048\\
CLINC150 & Laya & 0.717 & 0.105 & 0.913 & 0.160 & 0.714 & 0.080\\
MASSIVE & Jev & 0.380 & 0.075 & 0.245 & 0.036 & 0.455 & 0.061\\
MASSIVE & Laya & 0.753 & 0.258 & 0.884 & 0.046 & 0.841 & 0.124\\
\bottomrule
\end{tabular}
\end{table}

\begin{figure}[!htbp]
\centering
\includegraphics[width=\linewidth]{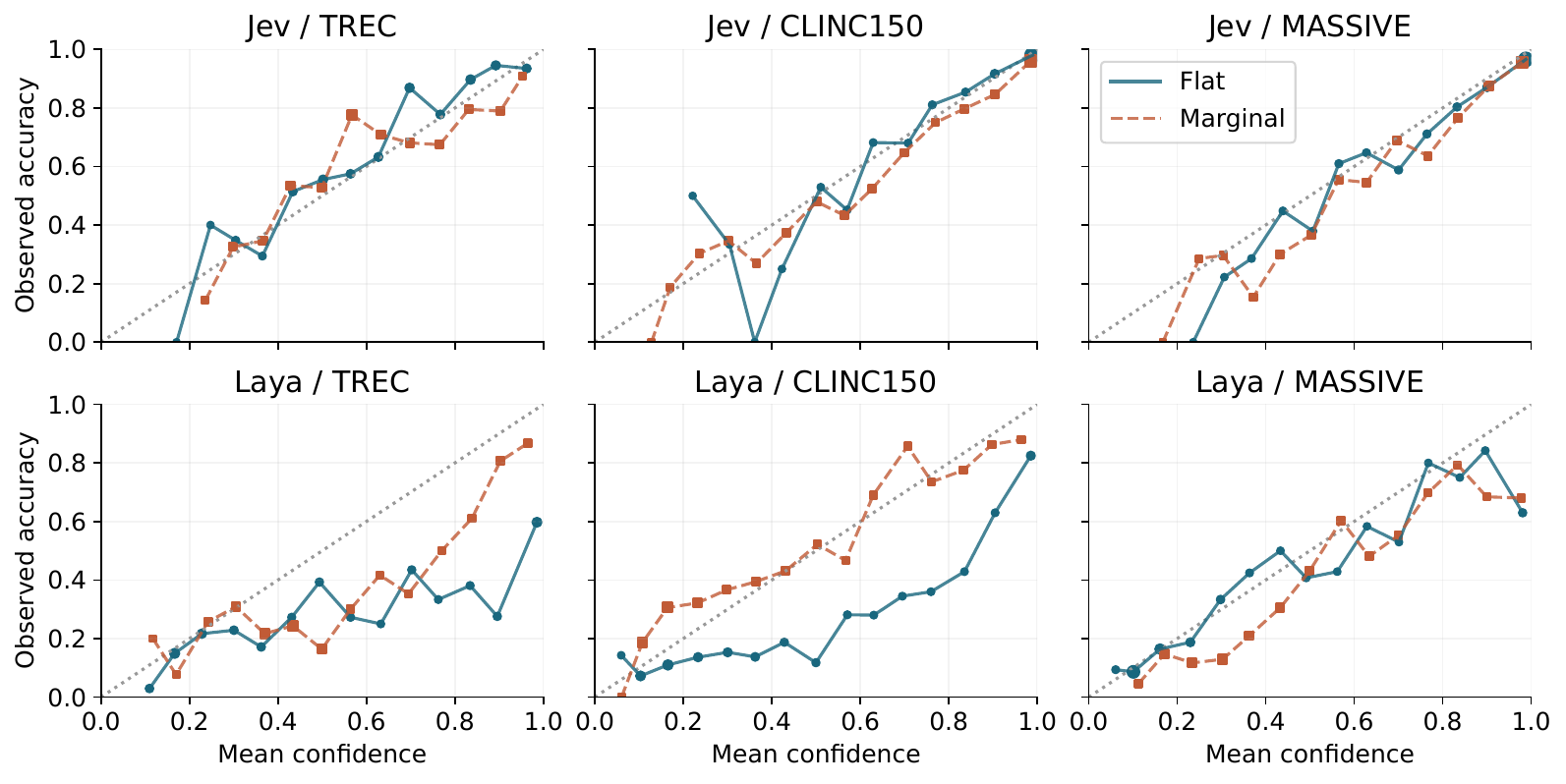}
\caption{Reliability diagrams for flat and marginal predictions, using 15 equal-width bins. Points show mean confidence and observed accuracy in nonempty bins; marker area increases with bin count. Lines connect the empirical bin summaries. Sparse bins have no uncertainty bars and should not be overinterpreted.}
\label{fig:reliability}
\end{figure}
Figure~\ref{fig:reliability} compares confidence with observed accuracy. Laya's flat MASSIVE predictions have relatively low ECE despite only 23.8\% accuracy, illustrating why calibration and predictive success need separate reporting.

The sign of the empirical marginal-minus-flat ECE difference is unchanged at 10, 15, and 20 bins for every model--dataset pair (Appendix~\ref{app:sensitivity}). This supports directional stability across the tested bin choices; statistical intervals retain the primary 15-bin specification.

\subsection{Coarse--fine and flat--marginal coherence}
\begin{figure}[!htbp]
\centering
\includegraphics[width=\linewidth]{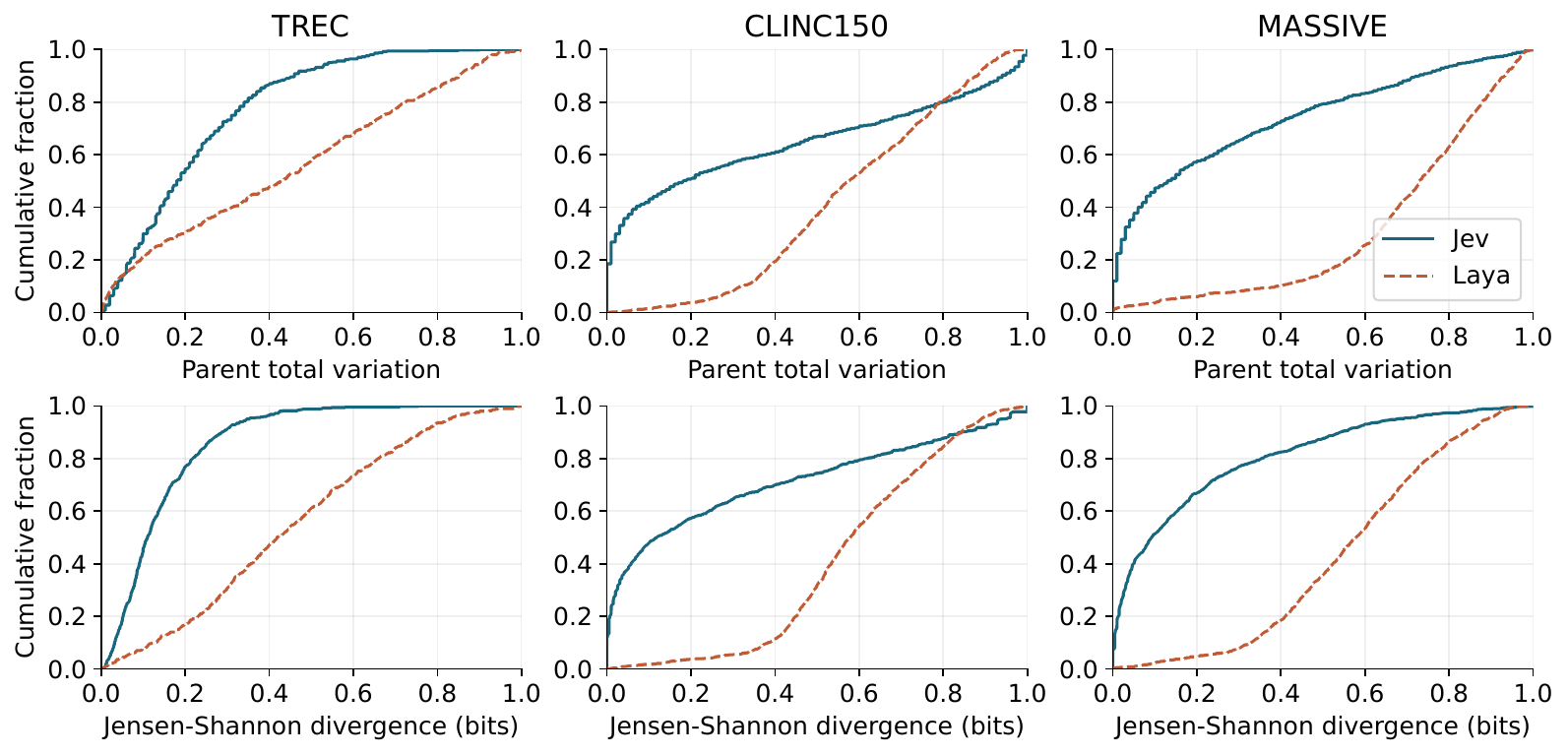}
\caption{Empirical cumulative distributions of parent TV (top) and flat--marginal JSD (bottom). A curve farther left places more examples at smaller discrepancies. These metrics measure disagreement between elicited distributions across all examples in each model--dataset group.}
\label{fig:coherence}
\end{figure}

Both systems assign substantially different probabilities across formulations, answering RQ1 and RQ2. Jev's mean parent TV ranges from 0.219 on TREC to 0.349 on CLINC150; Laya's ranges from 0.424 on TREC to 0.689 on MASSIVE (Table~\ref{tab:coherence}). Figure~\ref{fig:coherence} shows the variation across examples. Jev has smaller mean TV and JSD on every dataset, yet considerable disagreement remains even alongside its 91.0\% CLINC150 flat accuracy: mean parent TV is 0.349 and JSD is 0.281 bits.

\begin{table}[!htbp]
\centering\small
\setlength{\tabcolsep}{4pt}
\caption{Mean coherence discrepancies and flat--marginal label agreement. TV, drift, and JSD range from zero to one; JSD uses bits. CFCE and TV are algebraically redundant; compare TV across datasets.}
\label{tab:coherence}
\begin{tabular}{llrrrrr}
\toprule
Dataset & Model & CFCE & Parent TV & Gold drift & JSD & Agree (\%)\\
\midrule
TREC & Jev & 0.073 & 0.219 & 0.209 & 0.142 & 70.6\\
TREC & Laya & 0.141 & 0.424 & 0.223 & 0.438 & 36.0\\
CLINC150 & Jev & 0.070 & 0.349 & 0.314 & 0.281 & 69.3\\
CLINC150 & Laya & 0.117 & 0.587 & 0.297 & 0.591 & 12.3\\
MASSIVE & Jev & 0.029 & 0.257 & 0.228 & 0.188 & 75.7\\
MASSIVE & Laya & 0.077 & 0.689 & 0.219 & 0.572 & 18.4\\
\bottomrule
\end{tabular}
\end{table}

Flat and marginal predictions select the same label in 70.6\%, 69.3\%, and 75.7\% of Jev examples, compared with 36.0\%, 12.3\%, and 18.4\% for Laya. On CLINC150, Laya's lowest agreement accompanies its largest accuracy gain. Reconstruction changes many of its decisions and improves overall prediction, while substantial probability disagreement persists.

Gold-label drift illustrates why multiple metrics are necessary. Jev's drift is slightly larger than Laya's on CLINC150 (0.314 versus 0.297) and MASSIVE (0.228 versus 0.219), despite smaller full-distribution JSD. Drift concerns one annotated event, whereas JSD includes movement among all labels. Mean signed gold-probability change, an exploratory diagnostic, is negative for Jev ($-0.141$, $-0.279$, $-0.195$) and positive for Laya ($+0.097$, $+0.105$, $+0.103$), in TREC/CLINC150/MASSIVE order. Appendix~\ref{app:additional} gives intervals for the main coherence means.

\subsection{Sensitivity to option order and repeated requests}
Reordering the same choices frequently changes Laya's predictions (RQ4). Its flat labels change in 62.5\% of TREC, 61.7\% of CLINC150, and 76.7\% of MASSIVE permutation comparisons. Jev's corresponding rates are 20.0\%, 5.8\%, and 5.8\%. Laya's mean flat TV ranges from 0.450 to 0.554. These rates summarize 120 comparisons per dataset, formed from four reorderings of 30 distinct examples.

\begin{table}[!htbp]
\centering\small
\setlength{\tabcolsep}{4pt}
\caption{Option-order audit: mean TV and label-flip percentage against the main order. Each row uses 30 examples with four permutations (120 paired comparisons sharing 30 inputs).}
\label{tab:order}
\begin{tabular}{llrrrr}
\toprule
Dataset & Model & Flat TV & Flat flip (\%) & Marg. TV & Marg. flip (\%)\\
\midrule
TREC & Jev & 0.168 & 20.0 & 0.140 & 17.5\\
TREC & Laya & 0.491 & 62.5 & 0.221 & 39.2\\
CLINC150 & Jev & 0.062 & 5.8 & 0.150 & 19.2\\
CLINC150 & Laya & 0.554 & 61.7 & 0.368 & 41.7\\
MASSIVE & Jev & 0.069 & 5.8 & 0.097 & 14.2\\
MASSIVE & Laya & 0.450 & 76.7 & 0.247 & 25.8\\
\bottomrule
\end{tabular}
\end{table}

Marginalization reduces Laya's order-flip rates to 39.2\%, 41.7\%, and 25.8\%, leaving substantial sensitivity. Jev's marginal flip rates are 17.5\%, 19.2\%, and 14.2\%; the latter two exceed its flat rates. The effect of reconstruction on stability therefore depends on the system and dataset.

Jev also shows some variation when identical requests are repeated. Flat-label flips occur in 1 of 60 TREC comparisons and none of the CLINC150 or MASSIVE comparisons; marginal flips occur in 2, 5, and 5 comparisons (Table~\ref{tab:repeat}). Mean flat TV is 0.042, 0.017, and 0.016, respectively, below the order-audit means. These rounded-output comparisons suggest greater sensitivity to reordering in this audit, although identifying its causal mechanism requires further controls. Repeated-request stability remains untested for Laya.

\section{Error analysis}
\label{sec:errors}
\subsection{Which examples improve or deteriorate?}
\begin{figure}[!htbp]
\centering
\includegraphics[width=\linewidth]{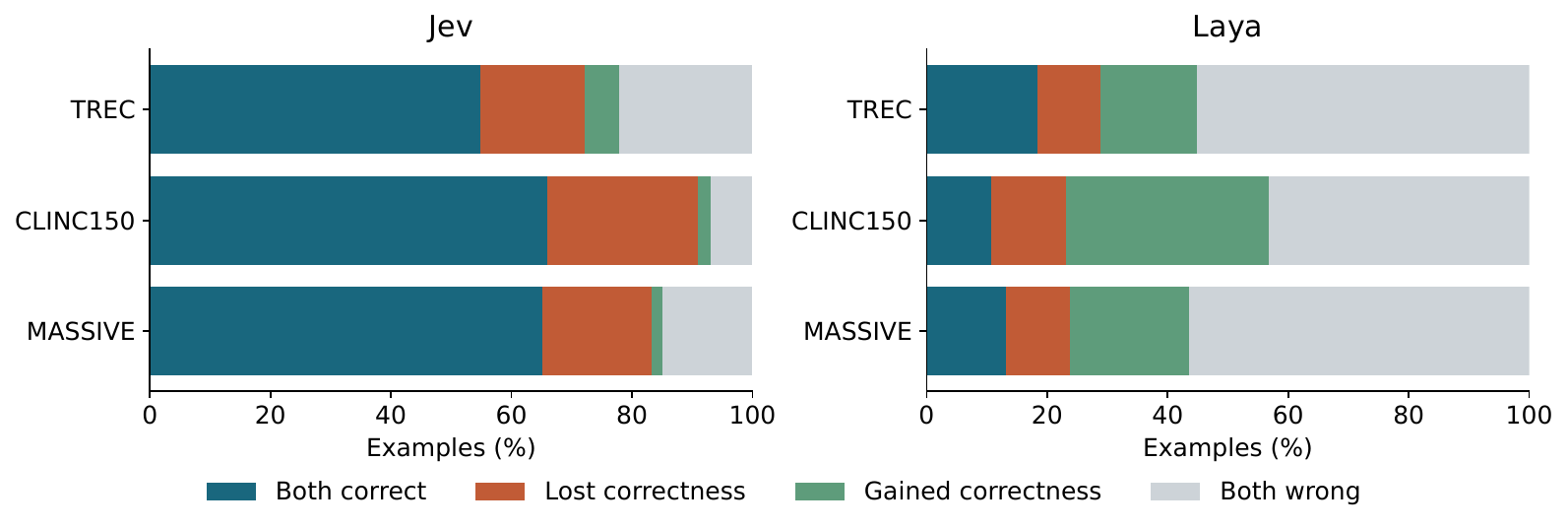}
\caption{Correctness transitions from flat to marginal classification. Segments partition all examples. The gained-minus-lost share equals the accuracy change; both-wrong cases may still change their predicted label.}
\label{fig:errors}
\end{figure}

Laya's accuracy gains include both corrections and new mistakes. On CLINC150, reconstruction corrects 337 examples but makes 124 previously correct examples wrong, giving a net gain of 213. More than half of its 231 originally correct decisions are lost. Jev shows the opposite balance: 250 lost decisions and 21 corrections give a net decrease of 229. On MASSIVE, Laya gains 198 and loses 106; on TREC it gains 80 and loses 52. Table~\ref{tab:errors} and Figure~\ref{fig:errors} show the complete breakdown.

\begin{table}[!htbp]
\centering\small
\setlength{\tabcolsep}{4pt}
\caption{Paired error accounting (counts). Both/lost/gained/neither refer to correctness under flat and marginal predictions and sum to N. Wrong parent plus within-parent errors plus hard-correct cases separately sum to N.}
\label{tab:errors}
\begin{tabular}{llrrrrrr}
\toprule
Dataset & Model & Both & Lost & Gained & Neither & Wrong parent & Within\\
\midrule
TREC & Jev & 274 & 87 & 28 & 111 & 117 & 98\\
CLINC150 & Jev & 660 & 250 & 21 & 69 & 317 & 37\\
MASSIVE & Jev & 651 & 182 & 18 & 149 & 271 & 70\\
TREC & Laya & 92 & 52 & 80 & 276 & 54 & 279\\
CLINC150 & Laya & 107 & 124 & 337 & 432 & 596 & 65\\
MASSIVE & Laya & 132 & 106 & 198 & 564 & 523 & 148\\
\bottomrule
\end{tabular}
\end{table}

Hard-routing accounting identifies different failure locations. On CLINC150, Jev has 317 wrong-parent decisions and only 37 wrong-child decisions among correctly routed examples. Conditional accuracy within the correctly selected parent is 94.6\%, but hard routing cannot recover the excluded branches. Laya has 596 wrong-parent and 65 within-parent errors there. Marginalization rescues 106 hard errors and loses one hard-correct case, explaining its 105-example improvement over hard routing.

For Laya on TREC, most hard-routing errors occur within the selected category. Only 54 of 500 parent decisions are wrong, while 279 examples have the correct parent and an incorrect child. Accuracy within correctly selected parents is 37.4\%. This locates the main difficulty at the fine-label stage of this workflow.

\subsection{Confidence and label-pair confusions}
At the flat-confidence threshold of 0.8, Jev has 13 errors among 169 selected TREC examples, 28 among 849 on CLINC150, and 43 among 735 on MASSIVE. Laya has 65 errors among 127 on TREC, 39 among 145 on CLINC150, and 27 among 89 on MASSIVE. A high-confidence error rate among selected examples differs from its fraction of all examples; Appendix~\ref{app:additional} provides denominators, accuracy, and subgroup JSD.

Label-pair counts identify specific distinctions to investigate. For Laya on TREC, gold \texttt{DESC:def} is predicted as \texttt{DESC:desc} 49 times under flat classification and 94 times under marginalization. Gold \texttt{HUM:ind} is mapped to \texttt{HUM:title} 16 and 30 times. This is consistent with difficulty distinguishing abbreviated labels, but no wording ablation establishes that expanded descriptions would repair it.

Jev's largest flat TREC confusion is \texttt{LOC:other} to \texttt{LOC:city} (14 examples). On CLINC150, all seven selected \texttt{reminder\_update} examples are flat-classified as \texttt{reminder}. On MASSIVE, its marginal classifier maps \texttt{play\_audiobook} to \texttt{audio\_volume\_other} 11 times. Correctness follows the dataset annotations, including distinctions between semantically close labels. These raw counts must be interpreted alongside the number of examples in each class.

\subsection{Illustrative cases and competing explanations}
The following cases are selected extremes under the rule in Section~\ref{sec:methods}. Probabilities refer to normalized recorded vectors; zeros and ones may reflect API rounding.

\paragraph{Broad-category interpretation can override a correct fine decision.}
For CLINC150 example \texttt{clinc:test:1118}, \emph{can you help me pay my electricity bill}, Jev assigns the correct fine label \texttt{pay\_bill} probability 1.000 under flat classification. The broad-category question favors \texttt{utility}, while the dataset places this intent under \texttt{banking}. Banking receives zero recorded probability, so reconstruction gives \texttt{pay\_bill} zero probability and selects \texttt{calculator}. The case illustrates a plausible ambiguity in the word \emph{utility}: its everyday meaning can conflict with the dataset's category assignment.

For \texttt{massive:test:9691}, \emph{open audio book history of rome}, Jev is correctly flat-classified as \texttt{play\_audiobook} with recorded probability 1.000. The coarse prediction is \texttt{audio}, while the gold parent is \texttt{play}, and the marginal prediction becomes \texttt{audio\_volume\_other}. On \texttt{trec:test:366}, a question about litmus-paper color in strong acid has flat gold probability 0.970, but marginal gold probability 0.076 after only 0.090 mass is assigned to the gold Entity parent.

\paragraph{Decomposition can recover task-relevant distinctions.}
For Laya's \texttt{trec:test:37}, \emph{How far is the service line from the net in tennis?}, flat classification selects \texttt{ENTY:sport}, whereas marginalization selects the correct \texttt{NUM:dist}. Gold probability rises from approximately 0.0001 to 0.951. For \texttt{clinc:test:48}, a seventy-dollar transfer request is flat-classified as \texttt{greeting} but correctly as \texttt{transfer} after marginalization; gold probability rises from 0.0015 to 0.997. For \texttt{massive:test:2638}, a news request is flat-classified as \texttt{iot\_wemo\_off} but marginal-classified correctly as \texttt{news\_query}.

Candidate-set size, instructions, ordering, and option-count calibration change together. Determining which factor explains each correction or failure requires separate experiments. Because these cases were selected for extreme changes, their frequency in the broader sample remains unspecified.

\section{Discussion}
\label{sec:discussion}
\subsection{Coherence belongs to the complete workflow}
RQ1 and RQ2 reveal disagreement at both the broad-category and fine-label levels. This can occur even when the direct classifier is correct. An application that substitutes a separately asked category probability for a sum of fine-label probabilities can therefore change the meaning of its numerical decisions. Checking coherence in the complete workflow is essential whenever those probabilities are combined.

\subsection{Accuracy gains coexist with probability disagreement}
RQ3 shows why predictive quality and coherence deserve separate reporting. Laya's reconstructed predictions improve accuracy and Brier score while remaining substantially different from its direct predictions. Jev loses accuracy under the same construction. The value of decomposition is therefore configuration-dependent, and probability agreement alone cannot determine which formulation is more useful.

Hard routing may suit an application that evaluates only one branch, at the risk of discarding the correct label after a parent error. Marginalization retains uncertainty across all branches and has higher observed accuracy here. Our evaluation collects every branch even when calculating hard predictions, so latency and cost trade-offs remain for future measurement.

\subsection{What the audits add}
RQ4 identifies choice order as an important source of variation, particularly for Laya. Jev is more stable under these permutations, although its reconstructed predictions can be less stable than its direct ones. Repeated-request variability adds a smaller source of variation in the Jev audit. Exact-tie and calibration-bin checks preserve the principal directions of the findings.

A useful next experiment would factorially vary parent wording, order, and calibration while holding other elements fixed. Another would compare independent coarse elicitation with deliberately coherent parent probabilities obtained by summing flat probabilities. Such constructed parent probabilities would provide an algebraically coherent reference. Both experiments remain future work.

\subsection{Implications for decision-system evaluation}
The practical lesson is to validate the formulation that an application will actually execute. First, compare direct and decomposed predictions on matched inputs. Second, keep conditional confidence distinct from unconditional fine-label confidence: a highly confident child of a low-probability parent need not be a high-probability overall answer. Third, report probability quality and coherence separately, because a more accurate reconstruction can still disagree strongly with the direct distribution. Finally, audit presentation changes before treating small score differences as stable. Deployment safety additionally requires application-specific validation.

\section{Limitations}
This exploratory study covers two configurations, three English closed-set datasets, and fixed label hierarchies. Scope narrowed after pilots and partial results. Instructions, choices, ordering, and calibration change together, limiting causal interpretation. Training exposure is unknown, and the systems differ in architecture and precision. Laya's expanded input budget and released temperature handling further qualify the comparison.

Approximately balanced sampling changes label frequencies; rare classes and duplicate texts limit resampling assumptions. Intervals describe the observed sample and are unadjusted for multiple comparisons. The order audit is small and reuses examples, repeated-request checks cover Jev only, and rounded outputs limit numerical precision. Methodology and appendices document these constraints, including the macro-F1 convention and high-cardinality setting.

\section{Conclusion}
This study examines whether probabilities returned for related classification questions agree across direct and hierarchical decisions. Across three datasets, Jev and Laya show substantial disagreement between directly estimated category probabilities and the sums of their fine-label probabilities. Combining category and within-category probabilities also produces distributions that differ from direct fine-label predictions. These differences have practical consequences: reconstruction reduces Jev's accuracy by up to 22.9 percentage points and improves Laya's by up to 21.3 points. The opposite effects show that the choice of decision formulation can substantially change a system's measured performance.

Accuracy gains also require careful interpretation. On MASSIVE, Laya gains 9.2 percentage points in accuracy while its expected calibration error rises from 0.046 to 0.124. Error analysis shows how mistakes at the category level can suppress a correct fine-label answer, while the order audit reveals additional sensitivity to the arrangement of choices. Together, these findings connect the four research questions: probability coherence, reconstructed predictions, confidence calibration, and order robustness describe complementary aspects of decision quality. Evaluating them together on matched examples provides a clearer basis for choosing and validating the decision workflow an application will use.

\clearpage
\bibliographystyle{unsrtnat}
\bibliography{references}

@INPROCEEDINGS{anwar2026dualalign,
  author={Anwar, Shuchismita and Rahman, Chowdhury Mofizur},
  booktitle={2026 5th International Conference on Electrical, Computer \& Telecommunication Engineering (ICECTE)},
  title={Dual-Align: Identity-Preserving Face Completion Through Topology-Consistent Latent Mapping and Boundary-Aware Refinement},
  year={2026},
  volume={},
  number={},
  pages={1-6},
  doi={10.1109/ICECTE69292.2026.11429227}
}

@article{gneiting,
  author={Tilmann Gneiting and Adrian E Raftery},
  title={Strictly Proper Scoring Rules, Prediction, and Estimation},
  journal={Journal of the American Statistical Association},
  volume={102},
  number={477},
  pages={359--378},
  year={2007},
  publisher={Taylor \& Francis},
  doi={10.1198/016214506000001437},
  URL={https://doi.org/10.1198/016214506000001437},
  eprint={https://doi.org/10.1198/016214506000001437}
}

@ARTICLE{lin,
  author={Lin, J.},
  journal={IEEE Transactions on Information Theory},
  title={Divergence measures based on the Shannon entropy},
  year={1991},
  volume={37},
  number={1},
  pages={145-151},
  doi={10.1109/18.61115}
}

@misc{typesafe,
  author={{TypeSafe AI}},
  title={{TypeSafe AI}},
  year={2026},
  url={https://typesafe.ai/},
  note={Accessed September 28, 2026}
}

@misc{laya,
  author={Mukkunnoth, Nandakishor},
  title={{Laya}: Non-autoregressive System 1 decision engine},
  year={n.d.},
  howpublished={GitHub repository},
  url={https://github.com/NandhaKishorM/laya},
  note={Accessed September 28, 2026}
}

@InProceedings{guo,
  title = 	 {On Calibration of Modern Neural Networks},
  author =       {Chuan Guo and Geoff Pleiss and Yu Sun and Kilian Q. Weinberger},
  booktitle = 	 {Proceedings of the 34th International Conference on Machine Learning},
  pages = 	 {1321--1330},
  year = 	 {2017},
  editor = 	 {Precup, Doina and Teh, Yee Whye},
  volume = 	 {70},
  series = 	 {Proceedings of Machine Learning Research},
  month = 	 {06--11 Aug},
  publisher =    {PMLR},
  url = 	 {https://proceedings.mlr.press/v70/guo17a.html}
}

@inproceedings{liroth,
    title = "Learning Question Classifiers",
    author = "Li, Xin  and
      Roth, Dan",
    booktitle = "{COLING} 2002: The 19th International Conference on Computational Linguistics",
    year = "2002",
    url = "https://aclanthology.org/C02-1150/"
}

@inproceedings{larson,
    title = "An Evaluation Dataset for Intent Classification and Out-of-Scope Prediction",
    author = "Larson, Stefan  and
      Mahendran, Anish  and
      Peper, Joseph J.  and
      Clarke, Christopher  and
      Lee, Andrew  and
      Hill, Parker  and
      Kummerfeld, Jonathan K.  and
      Leach, Kevin  and
      Laurenzano, Michael A.  and
      Tang, Lingjia  and
      Mars, Jason",
    editor = "Inui, Kentaro  and
      Jiang, Jing  and
      Ng, Vincent  and
      Wan, Xiaojun",
    booktitle = "Proceedings of the 2019 Conference on Empirical Methods in Natural Language Processing and the 9th International Joint Conference on Natural Language Processing (EMNLP-IJCNLP)",
    month = nov,
    year = "2019",
    address = "Hong Kong, China",
    publisher = "Association for Computational Linguistics",
    url = "https://aclanthology.org/D19-1131/",
    doi = "10.18653/v1/D19-1131",
    pages = "1311--1316"
}

@misc{fitzgerald,
      title={MASSIVE: A 1M-Example Multilingual Natural Language Understanding Dataset with 51 Typologically-Diverse Languages}, 
      author={Jack FitzGerald and Christopher Hench and Charith Peris and Scott Mackie and Kay Rottmann and Ana Sanchez and Aaron Nash and Liam Urbach and Vishesh Kakarala and Richa Singh and Swetha Ranganath and Laurie Crist and Misha Britan and Wouter Leeuwis and Gokhan Tur and Prem Natarajan},
      year={2022},
      eprint={2204.08582},
      archivePrefix={arXiv},
      primaryClass={cs.CL},
      url={https://arxiv.org/abs/2204.08582}, 
}

@article{betz,
    doi = {10.1371/journal.pone.0281372},
    author = {Betz, Gregor AND Richardson, Kyle},
    journal = {PLOS ONE},
    publisher = {Public Library of Science},
    title = {Probabilistic coherence, logical consistency, and Bayesian learning: Neural language models as epistemic agents},
    year = {2023},
    month = {02},
    volume = {18},
    url = {https://doi.org/10.1371/journal.pone.0281372},
    pages = {1-29},
    number = {2},

}

@misc{wolf,
      title={Partition, Prompt, Aggregate: Statistical Self-Consistency in Language Models}, 
      author={Patrik Wolf and Thomas Kleine Buening and Andreas Krause and Celestine Mendler-D{\"u}nner},
      year={2026},
      eprint={2607.15277},
      archivePrefix={arXiv},
      primaryClass={cs.CL},
      url={https://arxiv.org/abs/2607.15277}, 
}

@misc{zheng,
      title={Large Language Models Are Not Robust Multiple Choice Selectors}, 
      author={Chujie Zheng and Hao Zhou and Fandong Meng and Jie Zhou and Minlie Huang},
      year={2024},
      eprint={2309.03882},
      archivePrefix={arXiv},
      primaryClass={cs.CL},
      url={https://arxiv.org/abs/2309.03882}, 
}

@misc{sumit2026doesknowledgedistillationhurt,
      title={When Does Knowledge Distillation Hurt? Reliability-Aware Distillation for Low-Resource Language Summarization}, 
      author={Dipto Sumit and Ankan Kumar Roy Srizon and Sadia Khair Rodela and Atia Haque Asha and Mourchona Afrin and Niloy Farhan and Farig Sadeque},
      year={2026},
      eprint={2607.19956},
      archivePrefix={arXiv},
      primaryClass={cs.CL},
      url={https://arxiv.org/abs/2607.19956}, 
}

\clearpage
\appendix
\section{Additional uncertainty and audit results}
\label{app:additional}
These summaries use the same canonical records and normalization as the main text. Intervals concern means over the fixed stratified sample.
\begin{table}[!htbp]
\centering\small
\setlength{\tabcolsep}{4pt}
\caption{Mean parent TV and JSD with 95\% stratified bootstrap intervals.}
\label{tab:coherenceci}
\begin{tabular}{llrrrrrr}
\toprule
Dataset & Model & TV & Low & High & JSD & Low & High\\
\midrule
TREC & Jev & 0.219 & 0.208 & 0.230 & 0.142 & 0.134 & 0.149\\
TREC & Laya & 0.424 & 0.403 & 0.446 & 0.438 & 0.421 & 0.456\\
CLINC150 & Jev & 0.349 & 0.337 & 0.361 & 0.281 & 0.271 & 0.292\\
CLINC150 & Laya & 0.587 & 0.576 & 0.598 & 0.591 & 0.581 & 0.601\\
MASSIVE & Jev & 0.257 & 0.244 & 0.269 & 0.188 & 0.179 & 0.199\\
MASSIVE & Laya & 0.689 & 0.677 & 0.702 & 0.572 & 0.561 & 0.583\\
\bottomrule
\end{tabular}
\end{table}

\begin{table}[!htbp]
\centering\small
\setlength{\tabcolsep}{4pt}
\caption{Paired marginal-minus-flat changes in Brier and ECE with 95\% bootstrap intervals. Negative changes indicate lower loss or calibration error.}
\label{tab:calibrationci}
\begin{tabular}{llrr}
\toprule
Dataset & Model & Brier change & ECE change\\
\midrule
TREC & Jev & +0.159 [+0.135, +0.184] & +0.021 [-0.018, +0.057]\\
TREC & Laya & -0.186 [-0.228, -0.144] & -0.033 [-0.076, +0.010]\\
CLINC150 & Jev & +0.303 [+0.284, +0.323] & +0.028 [+0.010, +0.048]\\
CLINC150 & Laya & -0.199 [-0.228, -0.170] & -0.080 [-0.101, -0.042]\\
MASSIVE & Jev & +0.210 [+0.189, +0.231] & +0.026 [+0.005, +0.049]\\
MASSIVE & Laya & -0.043 [-0.069, -0.017] & +0.078 [+0.044, +0.096]\\
\bottomrule
\end{tabular}
\end{table}

\begin{table}[!htbp]
\centering\small
\setlength{\tabcolsep}{4pt}
\caption{Jev identical-request audit: 30 examples per dataset, each repeated twice (60 paired comparisons). These are descriptive summaries of rounded outputs.}
\label{tab:repeat}
\begin{tabular}{lrrrr}
\toprule
Dataset & Flat TV & Flat flip (\%) & Marg. TV & Marg. flip (\%)\\
\midrule
TREC & 0.042 & 1.7 & 0.040 & 3.3\\
CLINC150 & 0.017 & 0.0 & 0.041 & 8.3\\
MASSIVE & 0.016 & 0.0 & 0.030 & 8.3\\
\bottomrule
\end{tabular}
\end{table}

\begin{table}[!htbp]
\centering\small
\setlength{\tabcolsep}{4pt}
\caption{Exploratory flat-confidence and correctness subgroups. High confidence means maximum flat probability at least 0.8. JSD columns condition on flat correctness; subgroup membership is not randomized.}
\label{tab:subgroups}
\begin{tabular}{llrrrrr}
\toprule
Dataset & Model & High-conf. N & Errors & Acc. (\%) & JSD correct & JSD wrong\\
\midrule
TREC & Jev & 169 & 13 & 92.3 & 0.146 & 0.130\\
CLINC150 & Jev & 849 & 28 & 96.7 & 0.272 & 0.374\\
MASSIVE & Jev & 735 & 43 & 94.1 & 0.178 & 0.242\\
TREC & Laya & 127 & 65 & 48.8 & 0.292 & 0.498\\
CLINC150 & Laya & 145 & 39 & 73.1 & 0.496 & 0.619\\
MASSIVE & Laya & 89 & 27 & 69.7 & 0.464 & 0.606\\
\bottomrule
\end{tabular}
\end{table}

\clearpage
\begingroup
\makeatletter
\renewenvironment{table}[1][]{\par\medskip\noindent\begin{minipage}{\linewidth}\def\@captype{table}}{\end{minipage}\par\medskip}
\makeatother
\section{Post-hoc sensitivity checks}
\label{app:sensitivity}
Exact top-probability ties occur in 7 of Jev's 2,500 flat vectors and 4 of its marginal vectors, and in none of Laya's flat or marginal vectors. Even the most favorable arbitrary tie resolution for Jev's marginal predictions and the least favorable for its flat predictions cannot reverse the accuracy decreases (Table~\ref{tab:ties}). This check covers exact recorded ties; near-ties and unrounded outputs remain unassessed.
\begin{table}[!htbp]
\centering\small
\setlength{\tabcolsep}{4pt}
\caption{Post-hoc sensitivity to exact top-probability ties. Counts use recorded normalized vectors. Accuracy ranges (percent) cover arbitrary choices among tied maxima, holding all non-tied predictions fixed; these are deterministic tie-resolution bounds.}
\label{tab:ties}
\begin{tabular}{llrrrr}
\toprule
Dataset & Model & Flat ties & Flat range & Marg. ties & Marg. range\\
\midrule
TREC & Jev & 6 & 71.8--72.6 & 3 & 60.2--60.8\\
CLINC150 & Jev & 1 & 90.9--91.0 & 0 & 68.1--68.1\\
MASSIVE & Jev & 0 & 83.3--83.3 & 1 & 66.9--67.0\\
TREC & Laya & 0 & 28.8--28.8 & 0 & 34.4--34.4\\
CLINC150 & Laya & 0 & 23.1--23.1 & 0 & 44.4--44.4\\
MASSIVE & Laya & 0 & 23.8--23.8 & 0 & 33.0--33.0\\
\bottomrule
\end{tabular}
\end{table}

Table~\ref{tab:ecebins} reports the ECE bin-count check. The direction is stable across the three tested resolutions: ECE increases for Jev on each dataset, decreases for Laya on TREC and CLINC150, and increases for Laya on MASSIVE. Magnitudes remain bin-dependent, and the primary interval claims continue to use 15 bins only.
\begin{table}[!htbp]
\centering\small
\setlength{\tabcolsep}{4pt}
\caption{Post-hoc bin-count sensitivity: marginal-minus-flat top-label ECE with 10, 15, or 20 equal-width bins. Negative values indicate lower empirical ECE. These descriptive point estimates assess bin-count sensitivity.}
\label{tab:ecebins}
\begin{tabular}{llrrr}
\toprule
Dataset & Model & 10 bins & 15 bins & 20 bins\\
\midrule
TREC & Jev & +0.019 & +0.021 & +0.020\\
CLINC150 & Jev & +0.032 & +0.028 & +0.028\\
MASSIVE & Jev & +0.028 & +0.026 & +0.032\\
TREC & Laya & -0.038 & -0.033 & -0.038\\
CLINC150 & Laya & -0.080 & -0.080 & -0.073\\
MASSIVE & Laya & +0.084 & +0.078 & +0.079\\
\bottomrule
\end{tabular}
\end{table}

Removing the later identifier in each of MASSIVE's two duplicate-text pairs leaves 998 unique inputs. Marginal-minus-flat accuracy changes are then $-16.33$ percentage points for Jev and $+9.32$ for Laya, compared with $-16.4$ and $+9.2$ in the primary sample. The directions are unchanged. No duplicate texts occur in the selected TREC or CLINC150 samples. This descriptive check does not replace a cluster-aware uncertainty analysis.
\endgroup
\section{Reproducibility settings}
\label{app:implementation}
Seeds: primary 42, audit 43, bootstrap 44. Exact checkpoint, dataset, and primary-run identifiers are retained in the local source lock and manifests, alongside source URLs and dependency versions. API reproduction remains subject to request variability.

\paragraph{Candidate order.}
Identifiers follow lexicographically sorted labels: \texttt{L000}, \texttt{L001}, and so forth. Fine identifiers stay fixed across questions; parents have a separate list. NumPy shuffles choices using the first 16 hexadecimal characters of the SHA-256 digest of the serialized tuple containing seed 42, example identifier, and permutation index. Each conditional subset has its own permutation; both systems use this procedure.

\paragraph{Execution.}
Transient failures use bounded retries with backoff. Cached successful answers support resumption; completed examples require every question to succeed. Responses retain model, provider, configuration, example, question, permutation, and repeat identifiers. Scheduling changes are tracked separately from protocol changes. Validated probabilities use float64 arithmetic; records are saved atomically and rederived from cached answers. Integrity checks compare frozen inputs, original responses, normalized vectors, and derived records. Credentials stay local; latency and cost remain outside scope.
\clearpage
\section{Actual JSON inputs and outputs}
\label{app:json}
The following listings use the same real TREC example, \texttt{trec:test:37}, for both systems. The input is \emph{How far is the service line from the net in tennis ?}, including the original space before the question mark. Its annotated label is \texttt{NUM:dist} (distance), under \texttt{NUM} (Numeric).

\subsection{Jev: broad-category question}
The request body and response below are copied from the saved primary-run record. Every candidate and response field is included; indentation is adjusted for display. The six candidate identifiers refer to broad categories, with \texttt{L005} representing Numeric.

\begin{lstlisting}[style=studyjson,caption={Jev: actual broad-category request body.},label={lst:jevinput}]
{
  "model": "jev-1.13.0",
  "state": "How far is the service line from the net in tennis ?",
  "questions": {
    "classification": {
      "type": "choice",
      "instructions": "Classify the input into one of these broad categories.",
      "criteria": {
        "L005": "Numeric",
        "L003": "Human",
        "L001": "Description",
        "L002": "Entity",
        "L004": "Location",
        "L000": "Abbreviation"
      }
    }
  }
}
\end{lstlisting}
\begin{lstlisting}[style=studyjson,caption={Jev: complete saved response to Listing~\ref{lst:jevinput}.},label={lst:jevoutput}]
{
  "model": "jev-1.13.0",
  "answers": {
    "classification": {
      "type": "choice",
      "choice": "L005",
      "confidence": 0.64,
      "probabilities": {
        "L005": 0.71,
        "L000": 0.0,
        "L001": 0.18,
        "L002": 0.02,
        "L003": 0.01,
        "L004": 0.08
      }
    }
  },
  "usage": {
    "input_tokens": 393,
    "output_tokens": 80
  }
}
\end{lstlisting}

\clearpage
\subsection{Laya: the same broad-category question}
Laya runs through the local SDK. Listing~\ref{lst:layainput} serializes the \texttt{state} and \texttt{questions} arguments passed to \texttt{agent.predict}, reconstructed exactly from the cached task and evaluated backend code. Listing~\ref{lst:layaoutput} is the complete saved SDK response. Candidate order and wording match the Jev request.

\begin{lstlisting}[style=studyjson,caption={Laya: JSON representation of the actual SDK arguments.},label={lst:layainput}]
{
  "state": "How far is the service line from the net in tennis ?",
  "questions": {
    "classification": {
      "type": "choice",
      "instructions": "Classify the input into one of these broad categories.",
      "criteria": {
        "L005": "Numeric",
        "L003": "Human",
        "L001": "Description",
        "L002": "Entity",
        "L004": "Location",
        "L000": "Abbreviation"
      }
    }
  }
}
\end{lstlisting}
\begin{lstlisting}[style=studyjson,caption={Laya: complete saved response to Listing~\ref{lst:layainput}.},label={lst:layaoutput}]
{
  "model": "laya-rl-agent",
  "answers": {
    "classification": {
      "type": "choice",
      "choice": "L005",
      "probabilities": {
        "L005": 0.9659,
        "L003": 0.007,
        "L001": 0.0091,
        "L002": 0.0082,
        "L004": 0.0062,
        "L000": 0.0036
      },
      "confidence": 0.8872,
      "answer_confidence": 0.9659,
      "action": {
        "act_probability": 1.0
      }
    }
  },
  "usage": {
    "input_tokens": 63,
    "output_tokens": 0
  }
}
\end{lstlisting}

The study computes prediction confidence as the largest normalized entry in \texttt{probabilities}. The separate native \texttt{confidence}, \texttt{answer\_confidence}, and \texttt{action} fields above are preserved as returned.

\clearpage
\subsection{From the JSON answers to probability coherence}
The same input also receives a direct question over all 50 fine labels and one conditional question for each of the six broad categories. Below, the left listing gives the complete Numeric conditional question shared by both systems. The right listing selects fields from the corresponding native answers and the direct answers. Here \texttt{L040} means \texttt{NUM:dist}, and \texttt{L021} means \texttt{ENTY:sport}. The right-hand probabilities are explicitly selected entries from larger vectors.

\noindent\begin{minipage}[t]{0.49\linewidth}
\begin{lstlisting}[style=studyjson,basicstyle=\ttfamily\scriptsize,caption={Complete Numeric conditional question.},label={lst:conditional}]
{
  "type": "choice",
  "instructions": "Assuming the input belongs to Numeric, classify it into one of these specific categories.",
  "criteria": {
    "L049": "Numeric: weight",
    "L045": "Numeric: period",
    "L044": "Numeric: perc",
    "L048": "Numeric: volsize",
    "L046": "Numeric: speed",
    "L038": "Numeric: count",
    "L041": "Numeric: money",
    "L040": "Numeric: dist",
    "L039": "Numeric: date",
    "L047": "Numeric: temp",
    "L042": "Numeric: ord",
    "L043": "Numeric: other",
    "L037": "Numeric: code"
  }
}
\end{lstlisting}
\end{minipage}\hfill
\begin{minipage}[t]{0.49\linewidth}
\begin{lstlisting}[style=studyjson,basicstyle=\ttfamily\scriptsize,caption={Selected native answer fields. The outer model/question keys organize these excerpts.},label={lst:excerpts}]
{
  "jev": {
    "flat": {
      "choice": "L040",
      "probabilities": {
        "L040": 0.8200000000000001,
        "L021": 0.06999999999999999
      }
    },
    "conditional:NUM": {
      "choice": "L040",
      "probabilities": {
        "L040": 1.0
      }
    }
  },
  "laya": {
    "flat": {
      "choice": "L021",
      "probabilities": {
        "L040": 0.0001,
        "L021": 0.9439
      }
    },
    "conditional:NUM": {
      "choice": "L040",
      "probabilities": {
        "L040": 0.9843
      }
    }
  }
}
\end{lstlisting}
\end{minipage}

Each complete probability vector is normalized before analysis. Jev's flat vector sums to 0.99, so its distance probability is $0.82/0.99=0.828283$. Laya's flat vector sums to 0.9999, giving $0.0001/0.9999=0.00010001$. Laya's Numeric conditional vector also sums to 0.9999. Reconstruction therefore gives
\[
q_{\mathrm{Jev}}(\mathrm{distance})=0.71\times1.0=0.71,\qquad
q_{\mathrm{Laya}}(\mathrm{distance})
=0.9659\times\frac{0.9843}{0.9999}=0.950830.
\]

\begin{center}
\small
\begin{tabular}{@{}lrr@{}}
\toprule
Quantity for this input & Jev & Laya\\
\midrule
Direct distance probability, $p(\mathrm{distance})$ & 0.828283 & 0.000100\\
Numeric mass summed from direct labels, $a(\mathrm{NUM})$ & 0.828283 & 0.003300\\
Separately asked Numeric probability, $u(\mathrm{NUM})$ & 0.710000 & 0.965900\\
Reconstructed distance probability, $q(\mathrm{distance})$ & 0.710000 & 0.950830\\
Parent total variation (all six categories) & 0.168990 & 0.971600\\
Direct prediction & \texttt{NUM:dist} & \texttt{ENTY:sport}\\
Reconstructed prediction & \texttt{NUM:dist} & \texttt{NUM:dist}\\
\bottomrule
\end{tabular}
\end{center}
This example makes both coherence checks concrete. The asked Numeric probability differs from its summed fine-label mass, and the direct distance probability differs from its reconstructed value. Jev keeps the correct label despite these differences; Laya changes from sport to the correct distance label.

The example is the selected Laya TREC case from Section~\ref{sec:errors}.

\end{document}